\documentclass[runningheads,orivec]{llncs}
\usepackage[T1]{fontenc}
\usepackage{graphicx}
\usepackage{tikz}
\usetikzlibrary{positioning, arrows.meta, calc}

\usepackage{siunitx}
\usepackage{csquotes}
\usepackage{hyperref}

\usepackage{xcolor}

\usepackage[nolist]{acronym}
\begin{acronym}
    \acro{rl}[RL]{reinforcement learning}
    \acro{ppo}[PPO]{proximal policy optimization}
    \acro{drl}[DRL]{deep reinforcement learning}
    \acro{mdp}[MDP]{Markov decision process}
    \acro{mpc}[MPC]{model predictive control}
    \acro{1dof}[1-DoF]{one-degree-of-freedom}
    \acro{lqr}[LQR]{linear-quadratic regulator}
    \acro{lqi}[LQI]{linear-quadratic-integral regulator}
    \acro{lti}[LTI]{linear Time-Invariant}
    \acro{sb3}[SB3]{Stable Baselines3}
    \acro{trpo}[TRPO]{trust region policy optimization}
    \acro{icps}[ICPS]{industrial cyber-physical system}
    \acro{cps}[CPS]{cyber-physical system}
    \acro{pid}[PID]{proportional-integral-derivative}
    \acro{pinn}[PINN]{physics-informed neural network}
    \acro{cmdp}[CMDP]{constrained Markov decision process}
    \acroplural{cmdp}[CMDPs]{constrained Markov decision processes}
    \acro{cbf}[CBF]{control barrier function}
    \acro{rk4}[RK4]{4th order Runge-Kutta}
    \acro{ode}[ODE]{ordinary differential equation}
    \acro{relu}[ReLU]{rectified linear unit}
\end{acronym}

\begin{document}
\title{Integrating Physics-Informed Neural Networks for Safe Reinforcement Learning in a 1-DoF Helicopter System}
\titlerunning{PINNs for Safe RL}
%
\author{Georg Schäfer\inst{1,2} \and Jakob Rehrl\inst{1} \and Stefan Huber\inst{1}
}
\authorrunning{G. Schäfer et al.}
%
\institute{Josef Ressel Centre for Intelligent and Secure Industrial Automation, \\
Salzburg University of Applied Sciences, Salzburg, Austria \and
Paris Lodron University of Salzburg, Salzburg, Austria \\
\email{georg.schaefer@fh-salzburg.ac.at}}
\maketitle  
\begin{abstract}
Deep reinforcement learning (DRL) offers powerful control for industrial cyber-physical systems (ICPSs), but its \enquote{black-box} exploration risks violating strict hardware safety limits.
Typically, these constraints are managed through complex reward shaping.
In this work-in-progress paper, we embed a differentiable physics model directly into the proximal policy optimization (PPO) actor loss function.
By simulating short-horizon future trajectories during training, the policy is penalized for anticipated safety violations independent of the task-reward signal.
Evaluated on a simulated 1-degree-of-freedom helicopter testbed with strict pitch constraints, our physics-informed soft regularizations substantially reduce constraint violations while maintaining reliable target tracking.
\end{abstract}

\section{Introduction}

\Acp{icps} integrate physical processes with computational control, underpinning modern manufacturing and smart production.
\Ac{rl} has demonstrated significant success in optimizing these complex environments, offering data-driven solutions where mathematical modeling is challenging~\cite{kober2013reinforcement}.
However, a primary limitation of standard \ac{rl} is its \enquote{black-box} exploration strategy, which can cause irreversible hardware damage in real-world mechatronic systems.
The target system is the Quanser Aero 2\footnote{\url{https://www.quanser.com/products/aero-2/}}, a mechatronic laboratory testbed suitable for controlling the angle $\theta$ of a bar by actuating two rotors, see~\cite{SRHH24} for details. 
Traditionally, safety constraints in \ac{rl} are encoded purely within the reward definition, leading to complex and brittle reward shaping~\cite{garcia2015comprehensive}.
Other safe \ac{rl} paradigms rely on \acp{cmdp}~\cite{altman2021constrained}, \acp{cbf}~\cite{ames2019control}, or offline \ac{mpc} projections \cite{SRHH26b}.
To achieve anticipatory and safe control without massive computational overhead, we propose embedding a differentiable physics model directly into the loss function of the \ac{ppo} algorithm.
This approach conditions the actor network regarding safety constraints by penalizing predicted future violations via differentiable simulation \cite{raissi2019physics}.
The scientific value of this work-in-progress lies in demonstrating constraint satisfaction decoupled from the primary task-reward signal, providing a scalable mechanism for safe \ac{rl} deployment.

\section{Methodology}

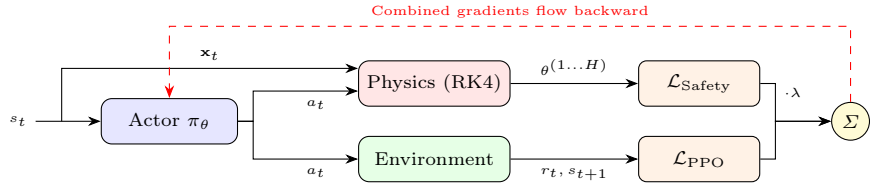
\begin{figure}[htbp]
    \centering
    \begin{tikzpicture}[
        >=Stealth,
        block/.style={draw, rectangle, rounded corners, minimum height=0.6cm, align=center, font=\scriptsize},
        actor/.style={block, fill=blue!10, minimum width=1.8cm},
        env/.style={block, fill=green!10, minimum width=2cm},
        pinn/.style={block, fill=red!10, minimum width=2cm},
        loss/.style={draw, rectangle, rounded corners=3pt, fill=orange!10, inner sep=2pt, font=\scriptsize, minimum width=1.6cm, minimum height=0.6cm, align=center},
        sum/.style={draw, circle, fill=yellow!20, inner sep=1pt, font=\scriptsize, minimum size=0.5cm},
        txt/.style={font=\tiny, align=center}
    ]

    \node[txt] (obs) at (0, 0) {$s_t$};
    \node[actor] (actor) at (2, 0) {Actor $\pi_\theta$};
    
    \node[pinn] (rk4) at (5.5, 0.5) {Physics (RK4)};
    \node[env] (env) at (5.5, -0.5) {Environment};
    
    \node[loss] (lsafety) at (9.0, 0.5) {$\mathcal{L}_{\text{Safety}}$};
    \node[loss] (lppo) at (9.0, -0.5) {$\mathcal{L}_{\text{PPO}}$};
    
    \node[sum] (ltotal) at (11.0, 0) {$\Sigma$};

    \draw[->] (obs) -- node[coordinate, pos=0.4] (split) {} (actor);
    
    \draw[->] (split) |- node[txt, above, pos=0.75] {$\mathbf{x}_t$} ([yshift=2mm]rk4.west);

    \draw[->] (actor.east) -- ++(0.2,0) |- node[txt, below, pos=0.8] {$a_t$} ([yshift=-1mm]rk4.west);
    \draw[->] (actor.east) -- ++(0.2,0) |- node[txt, below, pos=0.8] {$a_t$} (env.west);
    
    \draw[->] (rk4.east) -- node[txt, above] {$\theta^{(1...H)}$} (lsafety.west);
    \draw[->] (env.east) -- node[txt, below] {$r_t, s_{t+1}$} (lppo.west);

    \draw[->] (lsafety.east) -- ++(0.2,0) |- node[txt, above right, pos=0.3] {$\cdot \lambda$} (ltotal.west);
    \draw[->] (lppo.east) -- ++(0.2,0) |- (ltotal.west);

    \draw[->, dashed, red] (ltotal.north) -- ++(0, 1.0) -| node[txt, above, pos=0.25, text=red] {Combined gradients flow backward} (actor.north);

    \end{tikzpicture}
    \caption{Architecture overview: Actions are evaluated concurrently in the environment and the differentiable physics model. Both the task objective ($\mathcal{L}_{\text{PPO}}$) and the anticipatory safety penalty ($\mathcal{L}_{\text{Safety}}$, scaled by $\lambda$) combine into a total loss, whose gradients flow backward to update the policy.}
    \label{fig:pinn_architecture}
\end{figure}

Our base algorithm is \ac{ppo}~\cite{schulman2017proximal}, chosen for its stable performance and continuous action space capabilities.
To embed physical foresight, the system dynamics are modeled via known differential equations representing the Quanser Aero~2.
We integrate the \ac{pinn} via an auto-regressive unroll mechanism.
An overview of this dual-path architecture (evaluating actions concurrently in the environment for task reward and in the differentiable physics model for safety constraints) is illustrated in Fig.~\ref{fig:pinn_architecture}.

\begin{itemize}
    \item \textbf{Lookahead Horizon:} We project 0.3\,s into the future (3 steps at 10\,Hz).
    This balances the need to capture system inertia against the computational cost of long-term simulation.
    \item \textbf{System Dynamics vs. \ac{rl} Observation:} For every timestep $t$, we distinguish the agent's full observation state $s_t$ from the physical state $\mathbf{x_t}$ used for safety prediction.
    While the \ac{ppo} agent observes an augmented state \cite{SRHH26c}, the differentiable physics model isolates strictly the \ac{1dof} Quanser Aero~2 system state: $\mathbf{x} = [\theta, \dot{\theta}, \omega_0, \omega_1]^T$ (pitch angle, pitch velocity, and internal motor angular velocities).
    We use these core variables to initialize the non-linear continuous-time dynamics $\mathbf{\dot{x}}=f(\mathbf{x}, \mathbf{u})$ natively in PyTorch.
    During training, $\mathbf{x}_t$ is extracted directly from the simulator as privileged information, and the current policy is sampled for new actions using the predicted state as an input.
    \item \textbf{Numerical Integration:} We utilize a \ac{rk4} integration method, operating at 5 micro-steps per environment step, ensuring accurate trajectory simulation.
\end{itemize}

\textbf{Safety Constraint Definition:}
To rigorously test the mechanism, we impose an artificial pitch limit of $\pm 30^{\circ}$.
We simultaneously train the agent to track target signals of $\pm 40^{\circ}$.
While contradictory to standard operation, this extreme configuration effectively showcases the algorithm's ability to prioritize safety bounds over task-reward maximization.

\textbf{Loss Modification:}
The standard \ac{ppo} actor loss is augmented with a safety penalty derived from the differentiable physics model as $\mathcal{L} = \mathcal{L}_{\text{PPO}} + \lambda \cdot \mathcal{L}_{\text{Safety}}$, where $\lambda$ acts as a weighting factor.
Because the numerical \ac{rk4} integration is built entirely from trackable tensor operations, the computational graph records every simulation step.
Consequently, via the chain rule, gradients from anticipated safety violations flow backward through the unrolled differential equations into the actor network.
To compute $\mathcal{L}_{\text{Safety}}$, we unroll the system dynamics over the prediction horizon of $H=3$ steps.
Let $\theta^{(h)}$ denote the predicted pitch angle at horizon step $h$, simulated via the differentiable \ac{rk4} integrator.
The safety penalty acts as a soft regularizer, relying on a \ac{relu} to penalize the policy strictly when the absolute predicted pitch exceeds the predefined safety limit $\theta_{\text{max}} = 30^\circ$.
The normalized safety loss is defined as:

\begin{equation}
    \mathcal{L}_\text{Safety} = \frac{1}{H} \sum_{h=1}^H \max(0, \frac{|\theta^{(h)}|}{\theta_{\text{max}}}-1)
\end{equation}

Because this formulation is purely composed of differentiable tensor operations, the resulting gradients correctly map anticipated future constraint violations back to the actor network's current action distribution.

\section{Experimental Setup and Results}

We evaluate our approach on a simulated Quanser Aero~2 system in a \ac{1dof} configuration.
The underlying system dynamics and base \ac{rl} problem formulation have been established in previous studies \cite{SRHH24}, and the environment is accessed via a standard Gym API framework using Python-Simulink interfaces~\cite{SSRHH24}.
We tested three primary configurations, running 10 random seeds for each to ensure statistical significance: \emph{Naive Baseline} ($\lambda = 0$), \emph{PINN Balanced} ($\lambda = 150$), and \emph{PINN Over-penalized} ($\lambda = 1000$).
The penalty weight $\lambda$ was chosen empirically; a rigorous statistical evaluation of all 10 seeds and a $\lambda$ Pareto sweep are reserved for future work.

\textbf{Training Stability:}
All configurations successfully converged. We observed that the value loss naturally increases for higher $\lambda$ configurations due to the conflicting objectives (tracking a $40^{\circ}$ target vs.\ stopping at $30^{\circ}$), but the policy loss remained comparable across the board.

\begin{figure}[htbp]
    \centering
    \includegraphics[width=1.0\linewidth]{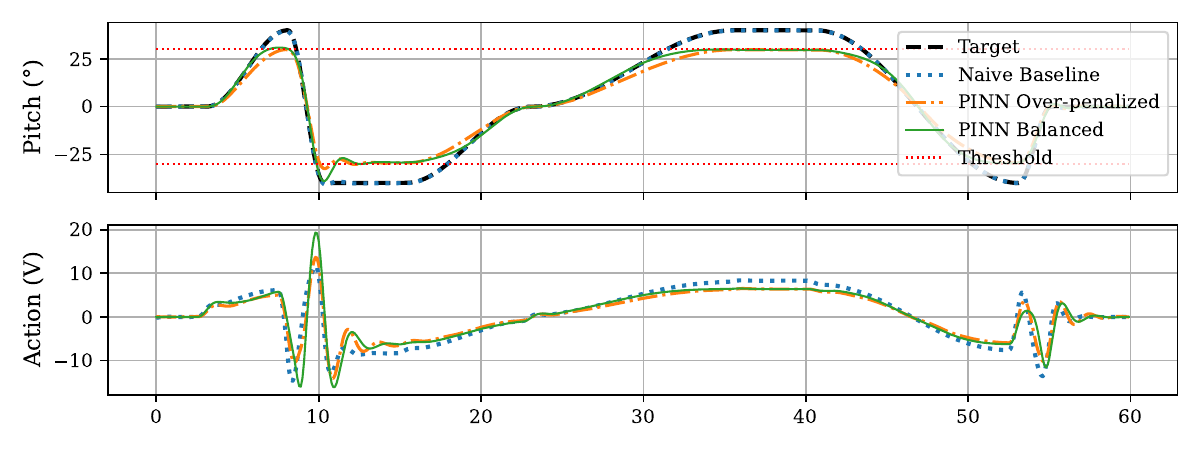}
    \caption{Simulation evaluation profile demonstrating the trade-off between target tracking and constraint satisfaction.
The \emph{Naive Baseline} ignores the $30^\circ$ safety threshold, the \emph{PINN Over-penalized} struggles to track the signal, and the \emph{PINN Balanced} successfully adheres to the constraints while maintaining strong tracking performance.}
    \label{fig:sim_eval}
\end{figure}

\textbf{Performance Trade-off and Safety:}
The evaluation results (see Fig.~\ref{fig:sim_eval}) highlight a clear trade-off between target tracking and constraint satisfaction.
The \emph{Naive Baseline} achieves high task reward by accurately tracking the $40^{\circ}$ target, but heavily violates the $30^{\circ}$ safety threshold.
The \emph{PINN Over-penalized} model exhibits very few safety violations but results in sluggish overall tracking, failing to adapt dynamically even when within safe regions.
The \emph{PINN Balanced} configuration demonstrates a suitable choice of $\lambda$.
It aggressively tracks the reference signal and largely learns to decelerate to respect the $30^\circ$ threshold.
Although brief transient violations can still occur during abrupt, extreme setpoint changes, it drastically reduces both predicted and actual pitch violations compared to the baseline.

\section{Conclusion and Future Work}

In this work-in-progress paper, we demonstrated that embedding differentiable physics models into the \ac{ppo} actor loss function provides a scalable and effective mechanism for soft anticipatory safety regularization in \acp{icps}.
This approach enables anticipatory constraint satisfaction and mitigates the need for complex, task-specific reward shaping.
Ongoing and future research will focus on the sim-to-real transfer of this \ac{pinn}-trained agent directly onto the physical Quanser Aero~2 hardware.
Beyond zero-shot transfer, we plan to conduct further fine-tuning directly on the real hardware to close the sim-to-real gap.
Because relying on a static \ac{ode} model during this phase could introduce systematic errors, this fine-tuning should be coupled with active parameter identification.
Building upon our recent work on parameter identification methodologies for the testbed \cite{LSRHH26}, we plan to dynamically update the parameters of the differentiable physics model to match observed real-world trajectories.
Finally, we aim to complement our anticipatory training loss with a reactive safety mechanism (such as a safety shield or \ac{cbf}) that acts as a real-time execution filter.
While the predictive loss smoothly guides the policy toward safe behaviors during training, a reactive architectural constraint would intervene directly during deployment, instantly overriding proposed actions if a forward-simulation predicts an imminent pitch violation.

\paragraph*{Acknowledgments.}{
Financial support for this study was provided by the Christian Doppler Research Association (CDG) through the Josef Ressel Centre for Intelligent and Secure Industrial Automation, the corresponding WISS Co-project of Land Salzburg, and by the European Interreg project BA0100172 AI4GREEN.
During the preparation of this work, the authors used AI tools for language editing and formatting assistance.
}

\bibliographystyle{splncs04}
\bibliography{references,jrcisia-published,jrcisia-pending}

@inproceedings{SRHH26c,
  title         = {{Anticipatory Reinforcement Learning for Trajectory Tracking}},
  author        = {Sch{\"a}fer, Georg and Rehrl, Jakob and Huber, Stefan and Hirlaender, Simon},
  booktitle     = {Database and Expert Systems Applications - DEXA 2026 Workshops},
  series        = {Communications in Computer and Information Science},
  publisher     = {Springer Cham},
  address       = {Graz, Austria},
  year          = 2026,
  note          = {submitted}
}

@inproceedings{LSRHH26,
  title         = {{Reinforcement Learning for Optimal Experiment Design in Parameter Identification of Mechatronic Systems}},
  author        = {Langschwert, Julian and Sch{\"a}fer, Georg and Rehrl, Jakob and Huber, Stefan and Hirlaender, Simon},
  booktitle     = {Database and Expert Systems Applications - DEXA 2026 Workshops},
  series        = {Communications in Computer and Information Science},
  publisher     = {Springer Cham},
  address       = {Graz, Austria},
  year          = 2026,
  note          = {submitted}
}

@inproceedings{SSRHH24,
  title         = {{Python-Based Reinforcement Learning on Simulink Models}},
  author        = {Sch{\"a}fer, Georg and Schirl, Maximilian and Rehrl, Jakob and Huber, Stefan and Hirlaender, Simon},
  year          = 2024,
  month         = 09,
  booktitle     = {11th International Conference on Soft Methods in Probability and Statistics (SMPS 2024)},
  address       = {Salzburg, Austria},
  doi           = {10.1007/978-3-031-65993-5_55}
}

@inproceedings{SRHH24,
  title         = {{Comparison of Model Predictive Control and Proximal Policy Optimization for a 1-DOF Helicopter System}},
  booktitle     = "2024 IEEE 22nd International Conference on Industrial Informatics (INDIN'24)",
  author        = {Sch{\"a}fer, Georg and Rehrl, Jakob and Huber, Stefan and Hirlaender, Simon},
  year          = 2024,
  month         = 08,
  publisher     = {{IEEE}},
  address       = {{Beijing, China}},
  doi           = {10.1109/INDIN58382.2024.10774357}
}

@inproceedings{SRHH26b,
  title         = {Safe Reinforcement Learning using Ideas from Model Predictive Control},
  author        = {Sch{\"a}fer, Georg and Rehrl, Jakob and Huber, Stefan and Hirlaender, Simon},
  booktitle     = {Computer {Aided} {Systems} {Theory} – {EUROCAST} 2026 – {Extended Abstracts}},
  year          = 2026,
  month         = 02,
  volume        = 20,
}

@book{altman2021constrained,
  title={Constrained Markov decision processes},
  author={Altman, Eitan},
  year={2021},
  publisher={Routledge}
}

@inproceedings{ames2019control,
  title={Control barrier functions: Theory and applications},
  author={Ames, Aaron D and Coogan, Samuel and Egerstedt, Magnus and Notomista, Gennaro and Sreenath, Koushil and Tabuada, Paulo},
  booktitle={2019 18th European control conference (ECC)},
  pages={3420--3431},
  year={2019},
  organization={Ieee}
}

@article{raissi2019physics,
  title={Physics-informed neural networks: A deep learning framework for solving forward and inverse problems involving nonlinear partial differential equations},
  author={Raissi, Maziar and Perdikaris, Paris and Karniadakis, George E},
  journal={Journal of Computational physics},
  volume={378},
  pages={686--707},
  year={2019},
  publisher={Elsevier}
}

@article{schulman2017proximal,
  title={Proximal policy optimization algorithms},
  author={Schulman, John and Wolski, Filip and Dhariwal, Prafulla and Radford, Alec and Klimov, Oleg},
  journal={arXiv preprint arXiv:1707.06347},
  year={2017}
}

@article{kober2013reinforcement,
  title={Reinforcement learning in robotics: A survey},
  author={Kober, Jens and Bagnell, J Andrew and Peters, Jan},
  journal={The International Journal of Robotics Research},
  volume={32},
  number={11},
  pages={1238--1274},
  year={2013},
  publisher={SAGE Publications Sage UK: London, England}
}

@article{garcia2015comprehensive,
  title={A comprehensive survey on safe reinforcement learning},
  author={Garc{\i}a, Javier and Fern{\'a}ndez, Fernando},
  journal={Journal of Machine Learning Research},
  volume={16},
  number={1},
  pages={1437--1480},
  year={2015}
}

\end{document}